\pdfoutput=1

\documentclass[11pt]{article}

\usepackage[final]{acl}
\usepackage{array}
\usepackage{tabularx}
\usepackage[most]{tcolorbox}
\usepackage{algorithm}
\usepackage{algorithmic}
\usepackage{pifont}
\usepackage{colortbl}
\usepackage{makecell}
\usepackage{enumitem} 
\usepackage{arydshln}
\usepackage{diagbox}
\usepackage{longtable}
\usepackage{rotating}
\usepackage{multirow}
\usepackage{pifont}
\usepackage{amsmath,amsfonts}
\usepackage{graphicx}
\usepackage{color}
\usepackage{pifont}
\usepackage{algorithm}
\usepackage{algorithmic}
\usepackage{verbatim}
\usepackage{booktabs}
\usepackage{bm}
\usepackage{colortbl}
\usepackage{soul}
\usepackage{color}
\usepackage{soul}
\usepackage{color}
\usepackage{subcaption}    

\usepackage{array}
\usepackage{graphicx}
\usepackage{booktabs}
\usepackage{amsmath}
\usepackage{makecell}
\usepackage{multirow}
\usepackage{bm}
\usepackage{pifont}
\usepackage{booktabs}
\usepackage{times}
\usepackage{latexsym}
\usepackage{tabularx}
\usepackage[T1]{fontenc}
\usepackage{multirow}
\usepackage{arydshln}

\usepackage{graphicx}
\usepackage[utf8]{inputenc}
\usepackage[T1]{fontenc}
\usepackage{CJKutf8}
\usepackage{microtype}
\usepackage{inconsolata}
\usepackage{times}
\usepackage{latexsym}
\usepackage[T1]{fontenc}
\usepackage{float}

\usepackage{microtype}

\usepackage{inconsolata}

\title{DMDTEval: An Evaluation and Analysis of LLMs \\ on Disambiguation in Multi-domain Translation
}

\author{Zhibo Man, Yuanmeng Chen, Yujie Zhang\textsuperscript{\dag}, Jinan Xu  \\
        Key Laboratory of Big Data \& Artificial Intelligence in Transportation, Ministry of Education\\
School of Computer Science and Technology, Beijing Jiaotong University, Beijing, 100044, China\\
        \texttt{\{zhiboman, yuanmengchen, yjzhang, jaxu\}@bjtu.edu.cn} \\}

\begin{document}
\maketitle
\begin{CJK*}{UTF8}{gkai}
\begin{abstract}
Currently, Large Language Models (LLMs) have achieved remarkable results in machine translation. However, their performance in multi-domain translation (MDT) is less satisfactory, the meanings of words can vary across different domains, highlighting the significant ambiguity inherent in MDT. Therefore, evaluating the disambiguation ability of LLMs in MDT, remains an open problem. To this end,  we present an  evaluation and analysis of
LLMs on \underline{\textbf{d}}isambiguation in \underline{\textbf{m}}ulti-\underline{\textbf{d}}omain \underline{\textbf{t}}ranslation (\textbf{DMDTEval}), our systematic evaluation framework 
consisting of three critical aspects: \textbf{(1)} we construct a translation test set with multi-domain ambiguous word annotation, \textbf{(2)} we curate a diverse set of disambiguation prompt strategies, and \textbf{(3)} we design precise disambiguation metrics, and study the efficacy of various prompt strategies on multiple state-of-the-art
LLMs. We conduct comprehensive experiments across 4 language pairs and 13 domains, our extensive experiments reveal a number of crucial findings that we believe will pave the way and also facilitate further research in the critical
area of improving the disambiguation of LLMs.

\end{abstract}

\section{Introduction}\label{111}
In recent years,  LLMs achieve the promising results in machine translation (MT) that demonstrate their
potential in practical applications
\cite{jiao2023chatgptgoodtranslatoryes,qian-etal-2024-large,feng2025mt}.
\begin{figure}[t]
    \centering
    \includegraphics[width=\linewidth]{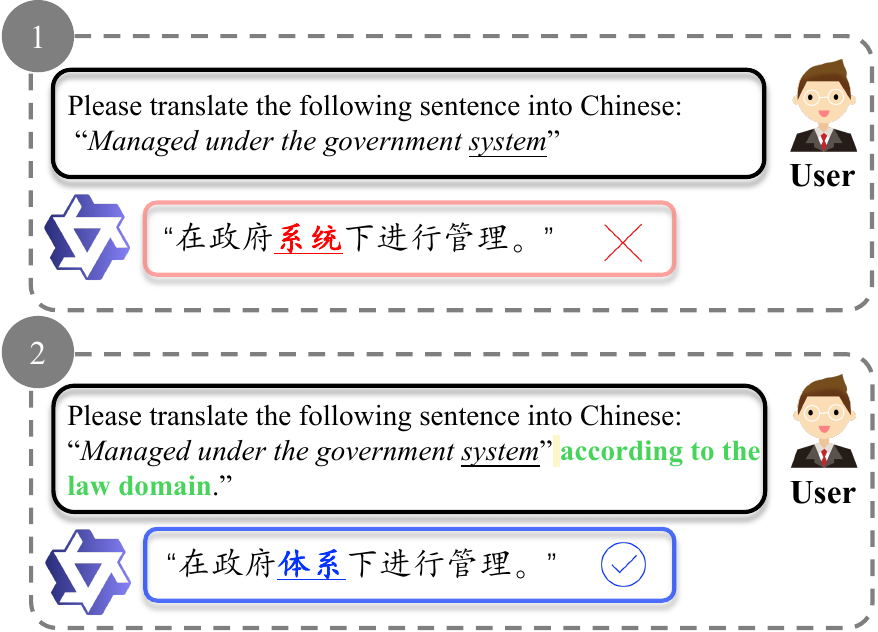}
    \caption{Two examples from the UM-Corpus English-Chinese test set. We prompt LLMs with domain label to disambiguate in Qwen-2.5-7B-Instruct. \textcolor{red}{{Red text}} represents for the ambiguity translation. \textcolor{blue}{{Blue text}} represents for the correct translation (\textit{hereinafter the same}).}
    \label{f1}
\end{figure}
However, LLMs perform unsatisfactorily in multi-domain translation (MDT) \cite{zheng2024fine,hu-etal-2024-large-language}.  
LLMs rely on extensive pre-training data, but multi-domain parallel corpora remain exceedingly scarce. This scarcity limits their translation capabilities and prevents them from effectively acquiring cross-domain knowledge, which leads to translation ambiguities.  
Figure \ref{f1}, Example \ding{172}, shows that directly using LLMs for translation causes word ambiguities. For example, the term “\underline{\textit{system}},” which refers to “\underline{体系}” (\textit{framework}), may be mistranslated as “\underline{系统}” (\textit{the literal translation of system}). This example illustrates that word ambiguity poses a key challenge for LLMs in MDT.

An intuitive solution is to directly prompt the LLMs to translate according to the specific domain \cite{hu-etal-2024-large-language}, and we find that this approach yields the correct translation. The translation of the term “\underline{\textit{system}}”  in the Law domain is accurate “\underline{体系}”, as shown in the  Figure \ref{f1}, Example \ding{173}. 
\textbf{The critical issue is how to effectively leverage domain information in prompt strategies to enhance the performance of LLMs.}

Regarding the above critical issue, previous work
mainly focuses on two key aspects: \textbf{(i)} \textbf{Multi-domain translation} \cite{jiang-etal-2020-multi-domain,man2023wdsrl,man2024ensemble}: these methods aim to enhance translation performance across different domains by incorporating sentence-level and word-level domain labels. Recently, some researchers have explored the performance of LLMs in MDT \cite{hu-etal-2024-large-language} and investigated fine-tuning LLMs using domain-specific parallel corpora \cite{hu-etal-2024-large-language,zheng2024fine}. \textbf{(ii)} \textbf{Disambiguation evaluation for translation}: \cite{campolungo2022dibimt,maheshwari2024dictdis,martelli2025dibimt}: these studies eluate the ability of models to handle and translate lexical ambiguities in general domains. 
The above-mentioned work provides feasible approaches for MDT under LLMs. However, three key research questions (\textbf{RQ}) remain unresolved in  MDT:
\begin{itemize}
\item \textbf{RQ1: \textit{How can we quantify the disambiguation ability of LLMs in MDT?}} Existing work (i) evaluates or fine-tunes MDT with LLMs, but does not address the role of key factors (\textit{i.e.,} ambiguity) that influence performance variation in MDT. Therefore, constructing an ambiguity dataset and designing evaluation metrics for ambiguity are crucial.

\item \textbf{RQ2: \textit{Can various prompting techniques help LLMs disambiguate in MDT?}}  
Figure \ref{f1}, Example \ding{173}, shows that the translation changes when the prompt includes domain information. This observation suggests that domain information influences the translation of LLMs. Therefore, we explore additional prompt strategies to determine how they affect the performance of LLMs in MDT.

\item \textbf{RQ3}: \textbf{\textit{What domain knowledge is essential for LLMs to achieve effective MDT?}}
Previous work (ii) mainly evaluates ambiguity in general domains.
In the MDT, the core research questions revolve around cross-domain word ambiguities and identifying which domain knowledge can be effectively leveraged under LLMs. 
\end{itemize}

To answer and explore the aforementioned questions, we  introduce an  evaluation and analysis of LLMs
on disambiguation in multi-domain translation for LLMs (DMDTEval) to tackle the challenges in MDT. 
\textbf{For RQ1}:  
We employ a word alignment tool to construct a multi-domain ambiguity vocabulary and manually annotate ambiguous words in the test set. Additionally, we design an evaluation metric to assess disambiguation ability in translation and compute the accuracy of ambiguous words being correctly translated.  
\textbf{For RQ2}:  
We design multiple disambiguation prompt strategiess to evaluate the translation performance of prominent LLMs across multiple domains. 
\textbf{For RQ3}: We conduct extensive experiments across four language pairs, with a particular focus on English-Chinese translation, providing a detailed and in-depth analysis along with key findings based on these experimental results.

To sum up, the main contributions of our work
can be summarized as follows:

\begin{itemize}

\item We construct an ambiguous word dataset specifically tailored for MDT. This dataset enables systematic evaluation of the disambiguation capabilities of LLMs. We will open-source it to support future research on enhancing the disambiguation performance of LLMs in MDT.
    
\item  We systematically explore various disambiguation prompt strategies, including zero-shot, chain-of-thought (CoT), few-shot, and reflection prompting, to evaluate MDT quality using 5 popular open-source LLMs.

\item We investigate the types of domain knowledge required by LLMs to evaluate translations across 4 language pairs and 13 domains, focusing on sentence-level and word-level domain knowledge, domain-specific examples, and domain discrimination capabilities.
\end{itemize}





\section{DMDTEval: Evaluation Framework}\label{s3}

\begin{figure*}[t]
    \centering
\includegraphics[width=\linewidth]{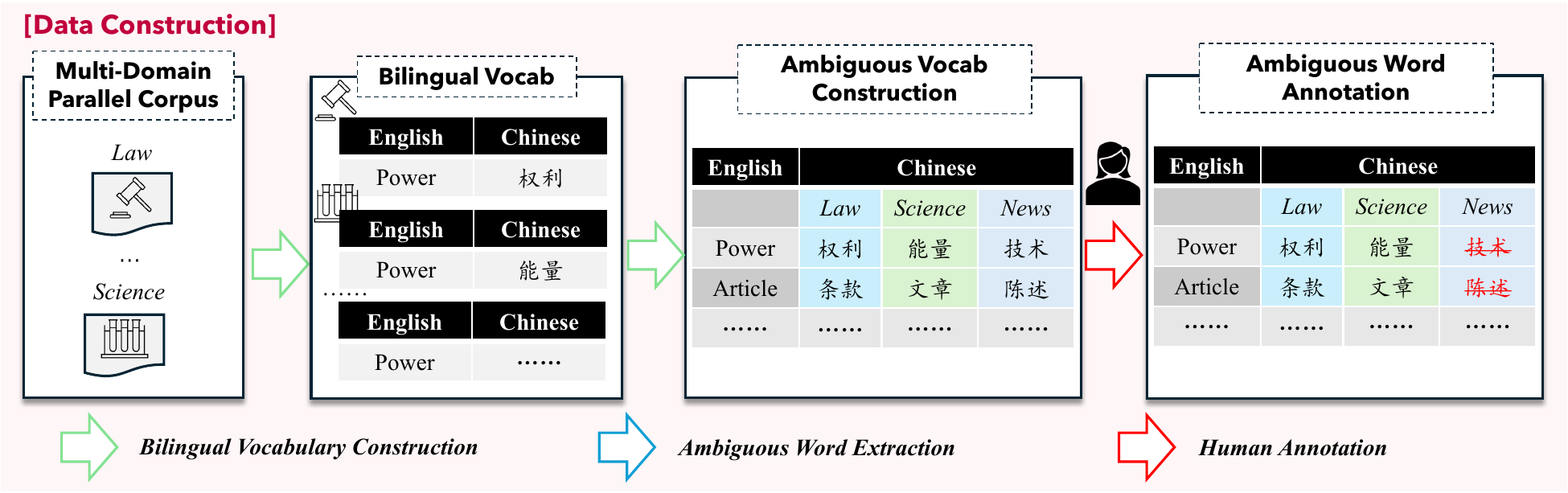}
    \caption{Ambiguous word test set construction annotation. This process consists of three steps.}
    \label{data2}
\end{figure*}

In our work, our goal includes \textbf{(1)} constructing a n ambiguous word test set (\S{\ref{s2.1}}). \textbf{(2)} evaluating the influence of domain information in LLMs' translation with different prompting (\S{\ref{s2.2}}). \textbf{(3)} and designing the metrics of word ambiguity (\S{\ref{s2.3}}).

\subsection{Data Construction}\label{s2.1}
\begin{figure}
    \centering
\includegraphics[width=\linewidth]{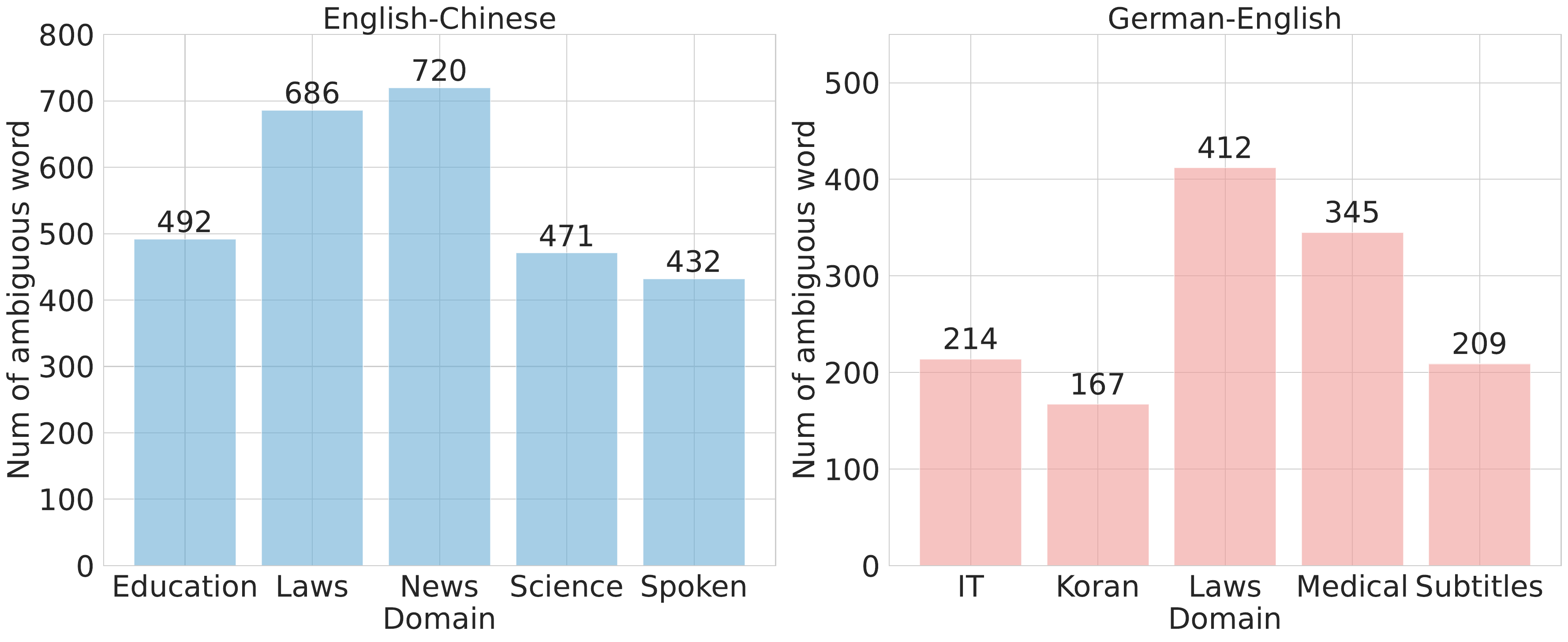}
    \caption{Statistics of ambiguous word in the test set. }
    \label{fig3}
\end{figure}



\vspace{0.5\baselineskip}
In this section, we aim to construct a multi-domain ambiguous word vocabulary to annotate the test set. 
Currently, the
publicly available test sets of domain-specific machine translation is scarce. We use the same dataset as in previous research \cite{man-etal-2024-icl,hu-etal-2024-large-language}, we mainly utilize two MDT test sets for ambiguous data set\footnote{Since this part of the data involves manual annotation, we primarily construct ambiguous data sets for English–Chinese and German–English, given our linguistic expertise in these language pairs. For Japanese–English and Korean–English, we utilize these data to evaluate overall translation quality.}: UM-Corpus\footnote{http://nlp2ct.cis.umac.mo/um-corpus/
} (English-Chinese), including five domains:  \textit{Education, Law, News Science, and Spoken} \cite{tian-etal-2014-um}, and OPUS\footnote{http://opus.nlpl.eu/} (German-English), including five domains: \textit{IT, Koran, Laws, Medical, and Subtitles}  \cite{aharoni-goldberg-2020-unsupervised}. The detailed statistic of these data sets in Appendix \ref{appdata}. We utilize the train set from these domains to obtain an ambiguity vocabulary, as shown in the Figure \ref{data2}.  
Our annotation processing consists of three steps: 

\vspace{0.5\baselineskip}

\noindent\textbf{Step 1: Bilingual Vocabulary Construction.}
In this step, we apply \textit{Awesome-Align}\footnote{https://github.com/neulab/awesome-align} \cite{dou2021word} to perform word alignment on multi-domain training corpora and extract bilingual word pairs. We then deduplicate and merge the bilingual vocabularies within each domain based on the source-language tokens. This process yields domain-specific bilingual lexicons, which include a substantial number of ambiguous words (\textit{e.g.,} “\textit{power}” → “权力” in Law domain, “\textit{power}” → “能量” in Science domain).

\vspace{0.5\baselineskip}

\noindent\textbf{Step 2: Ambiguous Vocabulary Construction.}
In this step, we construct a cross-domain ambiguous vocabulary based on the bilingual lexicons obtained in Step 1. For each domain, we initialize an empty set to store ambiguous word pairs. Then, for each bilingual pair in the domain-specific lexicon, we check whether the source word appears in other domains with different target-language translations. If such discrepancies are found, all corresponding translations are added to the ambiguous vocabulary set for that domain. This process results in a collection of domain-specific ambiguous vocabularies.

\vspace{0.5\baselineskip}

\noindent\textbf{Step 3: Human Annotation.}
Due to inevitable errors in word alignment, we manually refine the bilingual lexicons derived from the alignment process. In this step, we annotate the sentences in each domain's test set using the ambiguous vocabulary obtained in Step 2. Specifically, we identify and label instances of one-to-many source-language words that appear in the test set. The statistics of such ambiguous words are summarized in Figure~\ref{fig3}.

\vspace{0.5\baselineskip}

\noindent\textbf{Scoring of Alignment Quality.}
To evaluate the quality of the word alignments in ambiguous vocabulary construction, we randomly sample aligned word pairs from each domain and ask bilingual annotators to judge their correctness. Each pair is labeled as \texttt{correct}, \texttt{partially correct}, or \texttt{incorrect}. We calculate alignment accuracy as the proportion of correct alignments. Table~\ref{table77}  shows the results across domains in Appendix \ref{appdata}, highlighting the need for human annotation in Step 3.


\begin{figure*}[t]
    \centering
    \includegraphics[width=\linewidth]{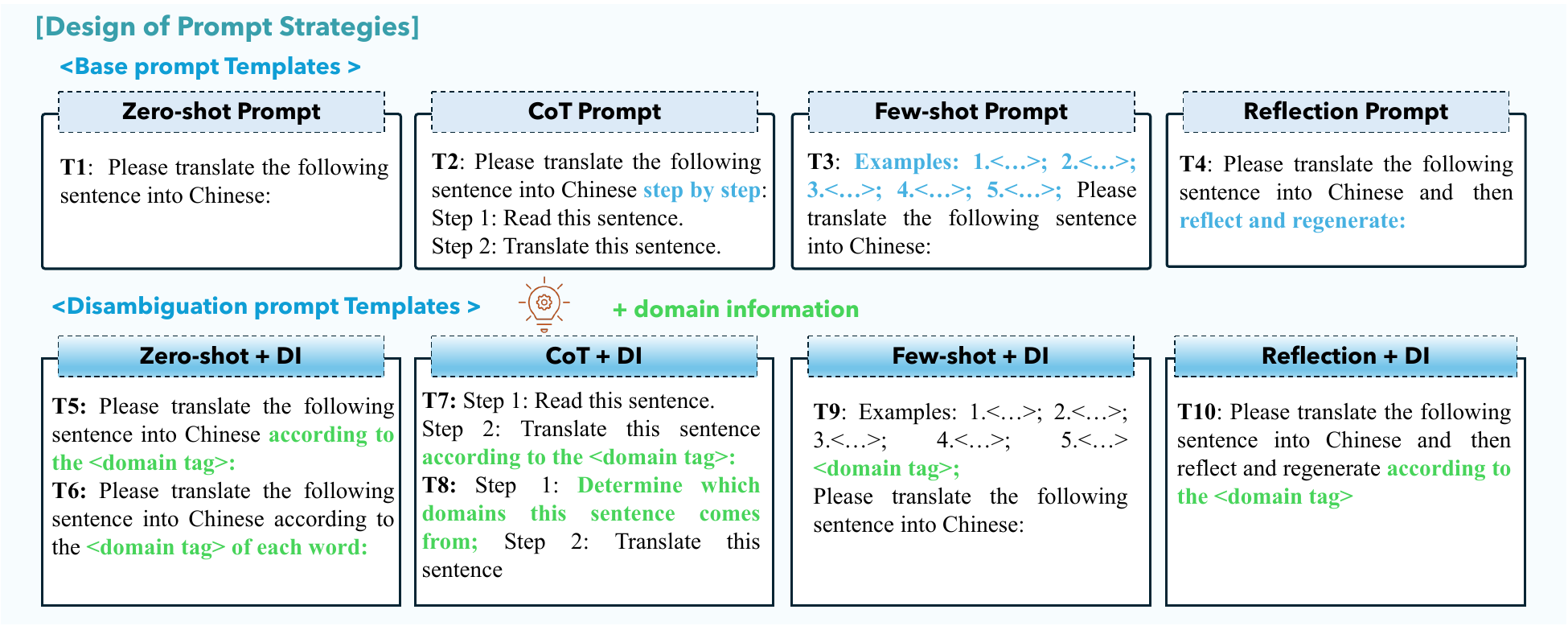}
    \caption{Design of Prompt Strategies. {Light blue text} represents for the specific information in each prompt strategy. {Light green text} represents for the specific information of disambiguation prompt strategies.}
    \label{f222}
\end{figure*}

\vspace{0.5\baselineskip}

\subsection{Design of Prompt Strategies}\label{s2.2}
In this section, we introduce the design of prompt templates, including both base prompt templates and disambiguation prompt templates, as shown in Figure \ref{f222}.
\vspace{0.5\baselineskip}

\noindent\textbf{Base Prompt Strategies.} Designing an effective prompt is the key to unlocking the translation capabilities of LLMs.
Specifically, we evaluate impact of different base prompt strategies, including: \textbf{(1)} \textbf{Zero-shot} directly asks LLM   to translate a source input into the target language \cite{liu-etal-2018-handling}. \textbf{(2)} \textbf{Chain-of-thought (CoT)} prompts  LLMs to reason about the input before generating an output \cite{wei2022chain}. \textbf{(3)} \textbf{Few-shot}: this prompt supplies an LLM with task-specific examples before querying it \cite{brown2020language}.
\textbf{(4)} \textbf{Reflection} \cite{shinn2024reflexion} further reflection on the generated translations yields new answers. 

\vspace{0.5\baselineskip}

\noindent\textbf{Disambiguation Prompt Strategies.} In this work, our prompt
includes 1) instructions to perform the task such as
“\textit{Please translate the following sentence into <target language>}” (\textit{i.e.,}, T1), and 2) domain  
information such as domain tag.
As shown in Figure \ref{f222}, our disambiguation prompts
strategies as following:

\textbf{(1) Zero-shot + domain information}: This strategies contain sentence-level and word-level:  1) Template 5 (T5): This template mainly utilize the domain information from the sentence-level domain tag base on the sentence-level MDT \cite{kobus-etal-2017-domain}. 1) Template 6 (T6): This template further utilize the domain information of each word base on the word-level MDT \cite{jiang-etal-2020-multi-domain}. We aim to evaluate whether fine-grained domain information can disambiguate and improve the capability of LLMs' understanding.

\textbf{(2) CoT + domain information}:
We also test whether
CoT prompting could improve LLMs’ performance
by utilizing reasoning-based steps for quality evaluation, Template 4 “\textit{Please translate the
following sentence into <target language> step by step: Step 1: read this sentence. Step 2: translate this sentence.}” Moreover, we design two disambiguation prompting by  devising Template 2: 1) {Template 7}: In this prompt, we give domain tag in step 2. This template further utilize domain information in reasoning；2) {Template 8}:  In this prompt, we ask LLMs to automatically discriminate which domain the source sentence comes from in step 1.

\textbf{(3) Few-shot + domain information}:
 We randomly retrieves 5-shot examples from
the training datastore and use  these
 examples for translation, this prompt is {Template 3}. To further integrate domain information, we add domain tags to each example, enhancing LLM's ability to perceive  domain as {Template 9}.

\textbf{(4) Reflection + domain information}:
Reflection encourages  LLMs to review and refine its  responses for improved accuracy and coherence \cite{shinn2024reflexion}. After reflecting on its initial output, the large model regenerates the translation as {Template 4}. We further enhance this process by incorporating domain information, encouraging the model to produce domain-specific translation results, as shown in Figure \ref{f222} {Template 10}. 

\subsection{Model Comparison and Selection}
\begin{figure}
    \centering
\includegraphics[width=\linewidth]{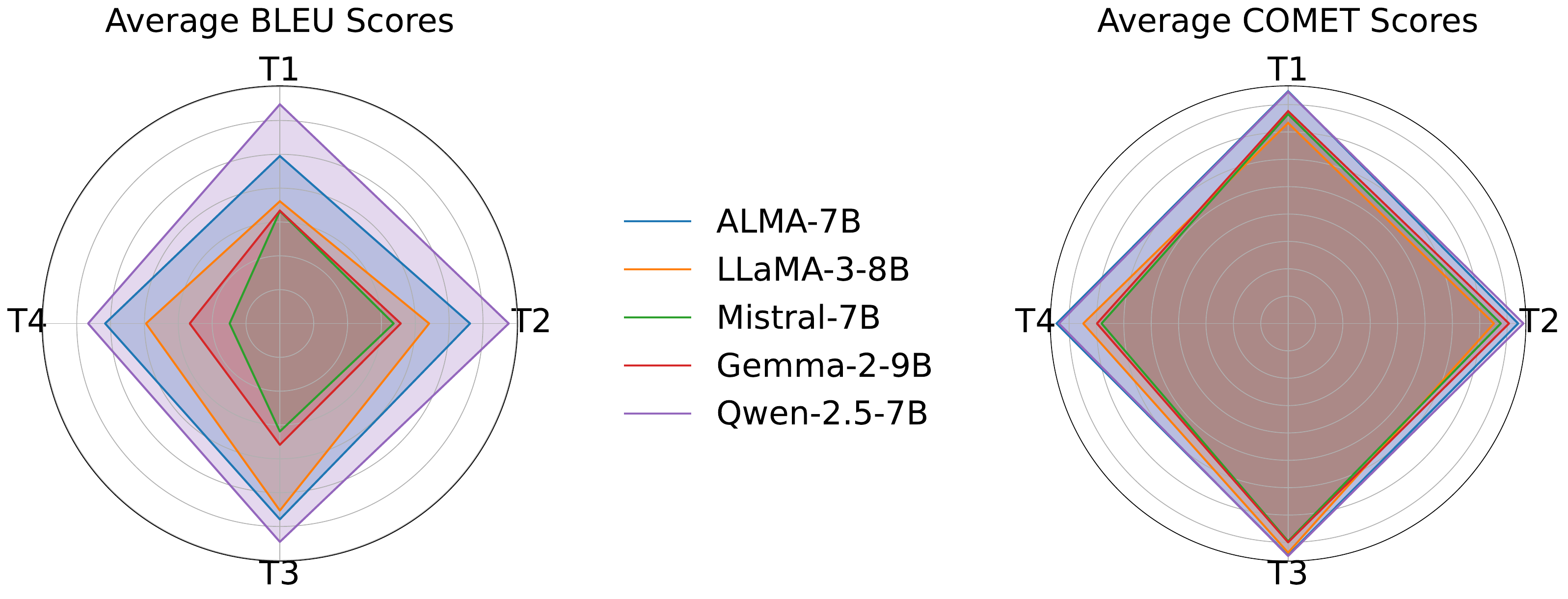}
    \caption{The comparison of different LLMs on the English-to-Chinese translation task (T1–T4) in terms of average BLEU and COMET scores.}
    \label{f5}
\end{figure}

In order to achieve more accurate and cost-effective replication, we are using the popular open-source model available at present. 
Our model selection can be divided into the following three categories: \textbf{(1)}\textbf{ Open-source}: we select LLama-3-8B \cite{grattafiori2024llama3herdmodels}, Mistral-7B \cite{jiang2024mixtral}, Gemma-2-9b \cite{team2024gemma}, and Qwen-2.5-7B  which
was specifically tested on a diverse set of 12 languages
and showed impressive multilingual capabilities \cite{bai2023qwentechnicalreport}. \textbf{(2)} \textbf{LLM-based translation model}: ALMA-7B fine-tuned in Llama-3-7B with translation instructions \cite{xu2024paradigmshiftmachinetranslation}.
 For all
5 selected models, we use the instruction-tuned
version, i.e., the chat model, for zero-shot, CoT and
few-shot inference. 
 As shown in the Figure \ref{f5}, Qwen-2.5-7B achieve the best performance on the English-Chinese MDT; therefore, we selected it as the base model for subsequent in-depth analysis in the section \ref{s3.2}.  \textbf{(3) NMT}:
NLLB \cite{nllb2022} is a multilingual translation model developed by Meta AI, supporting 200 languages. In addition, to demonstrate the performance on a larger-scale model, we also compare with Qwen-2.5-14B. The specific results are shown in Appendix \ref{appllm} Table \ref{t1-t4} and \ref{table10}.
 \begin{table*}[t]\small
\begin{center}
\begin{tabular}{llccccccc}
\Xhline{1pt}
\multirow{2}{*}{\textbf{Stragies}}&\multicolumn{6}{c}{\textbf{English-to-Chinese}}&\multirow{2}{*}{{\textbf{AVG}}}\\
\cline{3-7}
&&{\textbf{Education}}&{\textbf{Laws}}&{\textbf{News}}&{\textbf{Science}}&{\textbf{Spoken}}&\\
\hline
\multirow{5}{*}{\textbf{\makecell[l]{Zero-shot }}}&T1 & 33.14 / 88.10 & 50.82 / 88.94 & 30.04 / 84.51 & 28.76 / 84.82 & 19.20 / 77.00 & 32.39 / 84.67 \\
&\textcolor{orange}{T5}&33.46 / 88.21&51.39 / 89.20&30.36 / 84.92&28.78 / 86.13&20.89 / 77.46&\textbf{32.98} / \textbf{85.18}\\
&\scriptsize T5-T1&\textcolor{blue}{\scriptsize + 0.32 / 
+ 0.11} &\textcolor{blue}{\scriptsize + 0.57 / + 0.26}&\textcolor{blue}{\scriptsize + 0.32 / + 0.41}&\textcolor{blue}{\scriptsize + 0.02 / + 1.31}&\textcolor{blue}{\scriptsize + 1.69 / + 0.46}&\textcolor{blue}{\scriptsize + 0.59 / + 0.51} \\

&\textcolor{orange}{T6}&32.64 / 87.84&50.10 / 88.29&30.10 / 84.25&27.99 / 85.50&18.40 / 75.06&31.85 / 84.19\\
&\scriptsize T6-T1&\textcolor{red}{\scriptsize - 0.50 / - 0.26}&\textcolor{red}{\scriptsize - 0.72 / - 0.65}&\textcolor{blue}{\scriptsize + 0.06} \scriptsize  / \textcolor{red}{\scriptsize  - 0.26}&\textcolor{red}{\scriptsize - 0.77} \scriptsize / \textcolor{blue}{\scriptsize + 0.68}&\textcolor{red}{\scriptsize - 0.80} \scriptsize/ \textcolor{red}{\scriptsize - 1.94}&\textcolor{red}{\scriptsize - 0.54 / - 0.48} \\
\hdashline
\multirow{5}{*}{\textbf{\makecell[l]{CoT}}}&T2&34.02 / 88.06 & 51.19 / 89.60 & 30.51 / 84.91 & 28.82 / 85.91 & 22.45 / 79.31 & 33.40 / 85.56 \\
&\textcolor{orange}{T7}&34.50 / 88.09&52.09 / 90.15&31.00 / 85.15&28.97 / 86.05&23.47 / 80.88&\textbf{33.99} / \textbf{86.06}\\
&\scriptsize T7-T2&\textcolor{blue}{\scriptsize + 0.48 / + 0.03}& \textcolor{blue}{\scriptsize + 0.9 / + 0.55}&\textcolor{blue}{\scriptsize 	+ 0.49 / + 0.24}&\textcolor{blue}{\scriptsize + 0.15 / + 0.14}&\textcolor{blue}{\scriptsize 	+ 1.02 / + 1.57}&\textcolor{blue}{\scriptsize +	0.59 / + 0.50} \\
&\textcolor{orange}{T8}&33.56 / 88.22&50.39 / 88.79&30.15 / 84.88&28.95 / 86.05&22.02 / 79.01&32.61 / 85.79\\
&\scriptsize T8-T2&\textcolor{red}{\scriptsize  - 0.46 }\scriptsize /\textcolor{blue}{\scriptsize  + 0.16}&\textcolor{red}{\scriptsize - 0.80 / -0.81}&\textcolor{red}{\scriptsize -0.36 / -0.03}&\textcolor{blue}{\scriptsize + 0.13 / + 0.14}&\textcolor{red}{\scriptsize -0.43} \scriptsize / \textcolor{red}{\scriptsize - 0.30}&\textcolor{red}{\scriptsize - 0.79} \textcolor{blue}{\scriptsize / + 0.23} \\
\hdashline
\multirow{3}{*}{\textbf{\makecell[l]{Few-shot}}}&T3& 34.17 / 88.17 & 50.48 / 89.22 & 29.91 / 84.66 & 28.33 / 85.64 & 18.44 / 77.82 & \textbf{32.27} / \textbf{85.10} \\
&\textcolor{orange}{T9}&33.63 / 88.03&50.46 / 89.49&29.76 / 84.68&27.94 / 85.89&18.05 / 77.32&31.97 / 85.08 \\
&\scriptsize T9-T3&\textcolor{red}{\scriptsize - 0.54 / - 0.14}&\textcolor{red}{\scriptsize - 0.02} \scriptsize /\textcolor{blue}{\scriptsize  + 0.27}&\textcolor{red}{\scriptsize - 0.15} \scriptsize/ \textcolor{blue}{\scriptsize  + 0.02}&\textcolor{red}{\scriptsize - 0.39} / \textcolor{blue}{\scriptsize  + 0.25}&\textcolor{red}{\scriptsize   - 0.39 / - 0.50}&\textcolor{red}{\scriptsize  - 0.30} \scriptsize / \textcolor{red}{\scriptsize  - 0.02} \\
\hdashline
\multirow{3}{*}{\textbf{\makecell[l]{Reflection}}}&T4&26.75 / 86.06 & 47.77 / 87.76 & 26.16 / 82.71 & 25.90 / 84.03 & 17.01 / 76.05 & 28.72 / 83.32 \\
&\textcolor{orange}{T10}&32.80 / 87.83&50.61 / 89.16&30.24 / 84.57&28.60 / 85.68&22.20 / 79.40&\textbf{32.89} / \textbf{85.33}\\
&\scriptsize T10-T4&\textcolor{blue}{\scriptsize   + 6.05 / + 1.77}&\textcolor{blue}{\scriptsize   + 2.84 / + 1.40}&\textcolor{blue}{\scriptsize   + 4.08 / + 1.86}&\textcolor{blue}{\scriptsize   + 2.70 / + 1.65}&\textcolor{blue}{\scriptsize   + 	5.19 / + 3.35}&\textcolor{blue}{\scriptsize   + 4.17 / + 2.01} \\
\hline
\hline
\multirow{2}{*}{\textbf{Stragies}}&\multicolumn{6}{c}{\textbf{Chinese-to-English}}&\multirow{2}{*}{{\textbf{AVG}}}\\
\cline{3-7}
&&{\textbf{Education}}&{\textbf{Laws}}&{\textbf{News}}&{\textbf{Science}}&{\textbf{Spoken}}&\\
\hline
\multirow{5}{*}{\textbf{\makecell[l]{Zero-shot}}}&T1&
22.19 / 83.05&36.03 / 83.48&17.63 / 80.31&16.52 / 81.36&10.39 / 64.43&20.55 / 78.53\\
&\textcolor{orange}{T5}&
26.61 / 84.02&34.00 / 83.37&18.35 / 80.94&17.68 / 81.86&11.07 / 64.97&\textbf{21.94} / \textbf{79.83}\\
&\scriptsize T5-T1&\textcolor{blue}{\scriptsize  + 4.42 / + 0.97}&\textcolor{red}{\scriptsize   - 2.03 / - 0.11}&\textcolor{blue}{\scriptsize + 0.72 / + 0.63}&\textcolor{blue}{\scriptsize + 1.16 / + 0.50 }&\textcolor{blue}{\scriptsize + 0.68 / + 0.54}&\textcolor{blue}{\scriptsize + 1.39 / + 1.30}\\
&\textcolor{orange}{T6}&25.05 / 83.44&33.42 / 82.67&16.20 / 78.38&16.88 / 80.39&10.12 / 64.30&20.33 / 77.84\\
&\scriptsize T6-T1&\textcolor{blue}{\scriptsize + 2.86 / + 0.39}&\textcolor{red}{\scriptsize - 2.61 / - 0.81}&\textcolor{red}{\scriptsize - 1.43 / - 1.93}&\textcolor{blue}{\scriptsize + 0.36} \scriptsize/ \textcolor{red}{\scriptsize  - 0.97}&\textcolor{red}{\scriptsize  - 0.27 / - 0.13}&\textcolor{red}{\scriptsize  - 0.22 / - 0.69}\\
\hdashline
\multirow{5}{*}{\textbf{\makecell[l]{CoT}}}&T2&26.65 / 84.17&33.49 / 83.44&17.82 / 80.60&18.12 / 81.82&13.87 / 70.05&21.99 / 80.02\\
&\textcolor{orange}{T7}&26.63 / 84.91&34.46 / 83.72&18.08 / 80.29&18.82 / 81.86&14.59 / 71.97&\textbf{22.52} / \textbf{80.55}\\
&\scriptsize T7-T2&\textcolor{red}{\scriptsize - 0.02} \scriptsize / \textcolor{blue}{\scriptsize + 0.74}&\textcolor{blue}{\scriptsize + 0.97 / + 0.28}&\textcolor{blue}{\scriptsize+ 0.26 }\scriptsize / \textcolor{red}{\scriptsize - 0.31}&\textcolor{blue}{\scriptsize + 0.70 / + 0.04}&\textcolor{blue}{\scriptsize + 0.72 / + 1.92}&\textcolor{blue}{\scriptsize + 0.53 / + 0.53}\\
&\textcolor{orange}{T8}&
26.38 / 84.23&33.88 / 83.54&17.89 / 80.72&18.18 / 81.90&13.25 / 69.68&21.92 / 79.99\\
&\scriptsize T8-T2&\textcolor{red}{\scriptsize - 0.27} \scriptsize/ \textcolor{blue}{\scriptsize + 0.06}&\textcolor{blue}{\scriptsize + 0.39 / + 0.10}&\textcolor{blue}{\scriptsize + 0.07 / + 0.12}&\textcolor{blue}{\scriptsize + 0.06 / + 0.08}&\textcolor{red}{\scriptsize- 0.62 / - 0.37}&\textcolor{red}{\scriptsize - 0.07} \scriptsize/ \textcolor{red}{\scriptsize - 0.03}\\
\hdashline
\multirow{3}{*}{\textbf{\makecell[l]{Few-shot}}}&T3&26.36 / 84.05&32.08 / 83.44&18.67 / 80.69&18.26 / 81.80&12.30 / 68.91&21.53 / 79.78\\
&\textcolor{orange}{T9}&27.11 / 83.96&32.11 / 83.56&18.68 / 80.80&18.33 / 81.91&12.97 / 69.66&\textbf{21.84} / \textbf{79.98}\\
&\scriptsize T9-T3&\textcolor{blue}{\scriptsize + 0.75} \scriptsize/ \textcolor{red}{\scriptsize - 0.09}&\textcolor{blue}{\scriptsize + 0.03 / + 0.12}&\textcolor{blue}{\scriptsize + 0.01 / + 0.11}&\textcolor{blue}{\scriptsize + 0.07 / + 0.11}&\textcolor{blue}{\scriptsize + 0.67 / + 0.75}&\textcolor{blue}{\scriptsize + 0.31 / + 0.20}\\
\hdashline
\multirow{3}{*}{\textbf{\makecell[l]{Reflection}}}&T4&15.55 / 78.96&28.33 / 80.06&16.07 / 79.35&15.15 / 79.86&10.11 / 65.50&17.04 / 76.75\\
&\textcolor{orange}{T10}&24.33 / 83.69&34.25 / 83.77&17.72 / 80.68&16.84 / 81.87&13.52 / 66.30&\textbf{21.33} / \textbf{79.26}\\
&\scriptsize T10-T4&\textcolor{blue}{\scriptsize  + 8.78 / + 4.73}&\textcolor{blue}{\scriptsize + 5.92 / + 3.71}&\textcolor{blue}{\scriptsize + 1.65 / + 1.33}&\textcolor{blue}{\scriptsize + 1.69 / + 2.01}&\textcolor{blue}{\scriptsize + 3.41 / + 0.80}&\textcolor{blue}{\scriptsize + 4.29 / + 2.51}\\
\Xhline{1pt}
\end{tabular}
\end{center}
\caption{BLEU and COMET scores on the English-Chinese translation task for T1-T10 with Qwen-2.5-7B. We \textbf{bold} the best performance results in each strategy (\textit{hereinafter the same}). \textcolor{orange}{Orange} text stands for the disambiguation prompting templates. `` / '' represents for the ``BLEU / COMET''. \textcolor{blue}{Blue text} represents for the improvement and \textcolor{red}{red text} represents for the decrease (\textit{hereinafter the same}). }
\label{tablemain}
\end{table*}

\subsection{Evaluation Metrics}\label{s2.3}

\vspace{0.5\baselineskip}

\noindent\textbf{Translation Quality.}
We adopt two
widely-used metrics: SacreBLEU \cite{post2018call}, a n-gram matching-based metric, and the \texttt{wmt22-comet-da } model  is
used to generate the COMET\footnote{https://github.com/Unbabel/COMET} scores, the scope is 0-1, for convenience, we multiply the comet score by 100 in our experiments. In particular, we use
the paired bootstrap resampling methods \cite{koehn-2004-statistical} for the statistical significance test.

\vspace{0.5\baselineskip}

\noindent\textbf{Disambiguation Accuracy.}  
To evaluate the disambiguation ability of LLMs in MDT, we propose a metric based on the ambiguous vocabulary. Specifically, we identify all ambiguous source-language words in the test set and denote the total number of such instances as $n$. Among them, we count $m$ instances where the words are correctly translated according to their domain-specific meanings. We define disambiguation accuracy as $m/n$, which reflects how effectively an LLM resolves lexical ambiguity across domains. For example, in the science domain, the word \textit{``power''} should be translated as ``能量'' (\textit{energy}) rather than ``权力'' (\textit{authority}).

 \vspace{0.5\baselineskip}

\noindent
{\textbf{GPT-4o-mini Evaluator.}} Previous research \cite{qian-etal-2024-large} has shown that using GPT for translation quality evaluation is a feasible research approach. Therefore, we design a prompt to evaluate the disambiguation capability of LLMs using GPT-4o-mini\footnote{The specific prompt template for GPT-4o-mini Evaluator in Appendix \ref{appdis}}. 
\section{Evaluation Experiments}
In this section, we conduct an in-depth investigation of the three research questions (\textbf{RQs}) introduced in the Section \ref{111} through experiments on English-Chinese translation.
\vspace{0.5\baselineskip}

\noindent
\textbf{Evaluation and Training.} All our experiments were run using 1 × NVIDIA
V100 32G, for different LLM variants. We use vLLM\footnote{https://github.com/vllm-project/vllm} \cite{kwon2023efficient} to save inference time. 
 We keep the parameters consistent with those used in previous work \cite{qian-etal-2024-large}. For training, we use the Qwen-2.5-7B as base model for supervised fine-tuning base on the LLaMAFactory framework\footnote{
*https://github.com/hiyouga/LLaMA-Factory}. The Details of the training
procedure parameters are provided in Appendix \ref{appbb}.
\subsection{Main Results}\label{s3.2}
As shown in the Table \ref{tablemain}, incorporating domain information proves effective for the vast majority of prompt strategies, indicating that the disambiguation prompt strategies play a positive role. However, we also observed several noteworthy phenomena, which we further analyze in detail below:

\begin{tcolorbox}[colback=gray!5,colframe=black, arc=2mm,boxrule=0.5pt,width=\linewidth,boxsep=2pt, left=4pt, right=4pt,top=4pt, bottom=4pt] 
\small
\textbf{\textcolor{red}{Finding 1}}: \textbf{Adding domain information does not improve performance across all base prompt strategies on average score.
}
\end{tcolorbox}

\begin{table}[h]\small
    \centering
    \begin{tabular}{>{\raggedright\arraybackslash}p{1cm}|p{5.7cm}}
         \Xhline{1pt}
         \rowcolor{gray!12}
        \textbf{Domain} & \textbf{Education} \\
        \hline
        SRC & He washed his hands in a \textcolor{blue}{basin}. \\
        REF & 他在\textcolor{blue}{盆}里洗了手。 \\
        \hdashline
        T1 & 他用\textcolor{blue}{盆}洗了手。 \\
        T6 & 他用一个\textcolor{red}{盆子}洗了手。 \\
        \hline
        \hline
                 \rowcolor{gray!12}
        \textbf{Domain} & \textbf{News} \\
        \hline
        SRC & Is there a suicide \textcolor{blue}{contagion} on Wall Street?\\
        REF & 金融行业沦为自杀\textcolor{blue}{高发}行业？\\
        \hdashline
        T2 & 华尔街存在自杀\textcolor{blue}{高发}吗？\\
        T8 & 华尔街是否存在自杀\textcolor{red}{传播}？ \\
        \Xhline{1pt}
    \end{tabular}
    \caption{Three cases illustrate the phenomenon of decreased average scores for T6 and T8. }
    \label{table2}
\end{table}

\textbf{Analysis and Case Study for \textcolor{red}{Finding 1}.} As shown in Table \ref{table2}, for case 1, T6 adopts a word-based domain information translation strategy, focusing on lexical accuracy. As a result, the translations often exhibit clear word-to-word alignments, such as “a” → “一个” and “basin” → “盆子”. T8 automatically determines the domain of the sentence and then translates accordingly. This strategy may lead to translation errors (\textit{i.e.,} “contagion” → “传播”) if the domain is Economic domain.

\begin{tcolorbox}
[colback=gray!5,colframe=black, arc=2mm,boxrule=0.5pt,width=\linewidth,boxsep=2pt, left=4pt, right=4pt,top=4pt, bottom=4pt] 
\small
\textbf{\textcolor{red}{Finding 2}}: \textbf{Apart from Reflection, adding domain information to other strategies yields inconsistent improvements across domains, even in the best-performing approach on average, CoT with domain information.
}
\end{tcolorbox}

\textbf{Analysis  for \textcolor{red}{Finding 2}.} As shown in the Table \ref{tablemain},
for both English-to-Chinese and Chinese-to-English translation direction, CoT combined with domain information (\textit{i.e.,} T7) achieves the highest average BLEU and COMET scores, reaching ``33.99 / 86.06'' and ``22.52 / 80.55'', respectively. This indicates that the reasoning-based approach of LLMs can generate more accurate translations across multiple domains. Notably, the Reflection achieves consistent improvements across all domains when domain information is incorporated, suggesting that it effectively leverages domain knowledge during the reasoning process. In contrast, other strategies do not show consistent gains with domain information, which we hypothesize is due to their limited ability to enhance disambiguation performance.

\begin{tcolorbox}[colback=gray!5,colframe=black, arc=2mm,boxrule=0.5pt,width=\linewidth,boxsep=2pt, left=4pt, right=4pt,top=4pt, bottom=4pt] 
\small
\textbf{\textcolor{red}{Finding 3}}: \textbf{Different domains exhibit varying degrees of sensitivity to prompt templates.
}
\end{tcolorbox}
\begin{table}[h]\small
    \centering
    \begin{tabular}{>{\raggedright\arraybackslash}p{1cm}|p{5.7cm}}
         \Xhline{1pt}
         \rowcolor{gray!12}
        \textbf{Domain} & \textbf{Laws} \\
        \hline
        SRC & \textcolor{blue}{Chapter III} Fundamental Rights and Duties of the Residents\\
        REF & 
\textcolor{blue}{第三章} 居民的基本权利和义务\\
\hdashline
SRC & \textcolor{blue}{Chapter IX Supplementary Provisions}\\
 REF & 
\textcolor{blue}{第九章 附则}\\ 
        \Xhline{1pt}
    \end{tabular}
    \caption{Two cases illustrate the  specialized domain terminology and distinct textual styles. }
    \label{table4}
\end{table}

 \textbf{Analysis and Case Study for \textcolor{red}{Finding 3}.} For Chinese-to-Chinese, in the zero-shot setting, the Spoken domain sees notable gains from T1 to T5, with BLEU increasing by 1.69 and COMET by 0.46, while the Science domain under the CoT strategy shows minimal change from T2 to T7, with BLEU increasing by only 0.15 and COMET by 0.14. In contrast, the Reflection strategy, comparing T4 and T10, achieves consistent and substantial improvements across all domains. For example, in the Education domain, BLEU increases by 6.05 and COMET by 1.77, this is due to the presence of more prominent domain features, such as specialized terminology and distinct textual styles, as shown in Table \ref{table4}.
\begin{tcolorbox}[colback=gray!5,colframe=black, arc=2mm,boxrule=0.5pt,width=\linewidth,boxsep=2pt, left=4pt, right=4pt,top=4pt, bottom=4pt] 
\small
\textbf{\textcolor{red}{Finding 4}}: \textbf{In some domains, BLEU improves while COMET decreases, indicating that these metrics fail to adequately reflect the model’s ability to handle ambiguity in MDT.
}
\end{tcolorbox}
\begin{table}[h]\small
    \centering
    \begin{tabular}{>{\raggedright\arraybackslash}p{1cm}|p{5.7cm}}
         \Xhline{1pt}
         \rowcolor{gray!12}
        \textbf{Domain} & \textbf{News} \\
        \hline
        SRC & It’s clear he doesn't have any \textcolor{blue}{power}.\\
        REF & 他显然没有任何\textcolor{blue}{权力}。\\
        \hdashline
        T6 & 显然没有任何\textcolor{red}{力量} \\
        T8 & 他显然没有任何\textcolor{red}{权利}。 \\
        \Xhline{1pt}
    \end{tabular}
    \caption{One case illustrate the phenomenon of BLEU and COMET scores are not inconsistent for T6 and T8. }
    \label{table3}
\end{table}

 \noindent \textbf{Analysis and Case Study for \textcolor{red}{Finding 4}}. For the English-to-Chinese translation direction, we found that BLEU and COMET scores exhibit divergent trends in the News domain. To illustrate this phenomenon, we present a case where the English word “\textit{power}”—which can mean either “权势” (\textit{authority}) or “力量” (\textit{strength})-demonstrates lexical ambiguity, as shown in Table \ref{table3}. This ambiguity can lead to discrepancies in evaluation results when using BLEU and COMET, as each metric may favor different reference choices.

In summary, the aforementioned interesting findings further demonstrate the necessity of explicitly designing prompt templates to reveal and study the disambiguation capabilities of LLMs in MDT. In addition, we conducted experiments on multi-domain datasets for German–English, Japanese–English, and Korean–English translation directions\footnote{Japanese-English and Korean-English from the Flores-101 data set \cite{nllb-24}. We give a detailed data processing in the Appendix \ref{appdata}}. The overall trends are consistent with those observed in the English–Chinese experiments, which supports the validity and rationality of our proposed research motivation. Detailed results are provided in Appendix \ref{appyuyan}.

\subsection{Fine-tuning Resluts}
\begin{table*}[t]\small
   \begin{center}
\begin{tabular}{lclllllll}
\Xhline{1pt}
\multirow{2}{*}{\textbf{Strategies}} & \multicolumn{6}{c}{\textbf{English-to-Chinese}} & \multirow{2}{*}{\textbf{AVG}} \\
\cline{3-7}
&& \textbf{Education} & \textbf{Laws} & \textbf{News} & \textbf{Science} & \textbf{Spoken} & \\
\hline
\multirow{3}{*}{\textbf{\makecell[l]{Zero-shot}}}
&T1 & 39.68 & 40.85 & 46.89 & 36.98 & 42.88 & 41.46 \\
&\textcolor{orange}{T5} 
& 42.56\textsubscript{\textcolor{blue}{+2.88}} 
& 44.96\textsubscript{\textcolor{blue}{+4.11}} 
& 47.69\textsubscript{\textcolor{blue}{+0.80}} 
& 44.12\textsubscript{\textcolor{blue}{+7.14}} 
& 43.65\textsubscript{\textcolor{blue}{+0.77}} 
& \textbf{44.60}\textsubscript{\textcolor{blue}{+3.14}} \\
&\textcolor{orange}{T6}
&36.36\textsubscript{\textcolor{red}{-3.32}}
&38.19\textsubscript{\textcolor{red}{-2.66}}
&45.11\textsubscript{\textcolor{red}{-1.78}}
&35.20\textsubscript{\textcolor{red}{-1.78}}
&40.56\textsubscript{\textcolor{red}{-2.32}}
&39.08\textsubscript{\textcolor{red}{-2.38}} \\
\hdashline
\multirow{3}{*}{\textbf{\makecell[l]{CoT}}}
&T2&44.60&45.97&47.22&45.14&44.90&45.57\\
&\textcolor{orange}{T7}
&45.04\textsubscript{\textcolor{blue}{+0.44}}
&46.98\textsubscript{\textcolor{blue}{+1.01}}
&48.27\textsubscript{\textcolor{blue}{+1.05}}
&46.05\textsubscript{\textcolor{blue}{+0.91}}
&45.98\textsubscript{\textcolor{blue}{+1.08}}
&\textbf{46.46}\textsubscript{\textcolor{blue}{+0.89}} \\
&\textcolor{orange}{T8}
&36.25\textsubscript{\textcolor{red}{-8.35}}
&38.69\textsubscript{\textcolor{red}{-7.28}}
&38.94\textsubscript{\textcolor{red}{-8.28}}
&30.10\textsubscript{\textcolor{red}{-15.04}}
&39.50\textsubscript{\textcolor{red}{-5.40}}
&36.70\textsubscript{\textcolor{red}{-8.87}} \\
\hdashline
\multirow{2}{*}{\textbf{\makecell[l]{Few-shot}}}
&T3 & 40.55 & 41.60 & 48.11 & 37.25 & 43.00 & 42.10 \\
&\textcolor{orange}{T9} 
&34.58\textsubscript{\textcolor{red}{-5.97}} 
&39.26\textsubscript{\textcolor{red}{-2.34}} 
&42.30\textsubscript{\textcolor{red}{-5.81}} 
&35.44\textsubscript{\textcolor{red}{-1.81}} 
&40.87\textsubscript{\textcolor{red}{-2.13}} 
&38.49\textsubscript{\textcolor{red}{-3.61}} \\
\hdashline
\multirow{2}{*}{\textbf{\makecell[l]{Reflection}}}
&T4 & 43.56 & 42.07 & 47.85 & 39.60 & 44.09 & 43.43 \\
&\textcolor{orange}{T10} 
&45.02\textsubscript{\textcolor{blue}{+1.46}} 
&43.05\textsubscript{\textcolor{blue}{+0.98}} 
&47.28\textsubscript{\textcolor{red}{-0.57}} 
&40.15\textsubscript{\textcolor{blue}{+0.55}} 
&45.60\textsubscript{\textcolor{blue}{+1.51}} 
&\textbf{44.22}\textsubscript{\textcolor{blue}{+0.79}} \\
\Xhline{1pt}
\end{tabular}
\end{center}
\caption{Disambiguation accuracy scores (\%) on English-to-Chinese translation task for T1--T10 with Qwen-2.5-7B.}
\label{table5}
\end{table*}

\begin{figure}
    \centering
    \includegraphics[width=\linewidth]{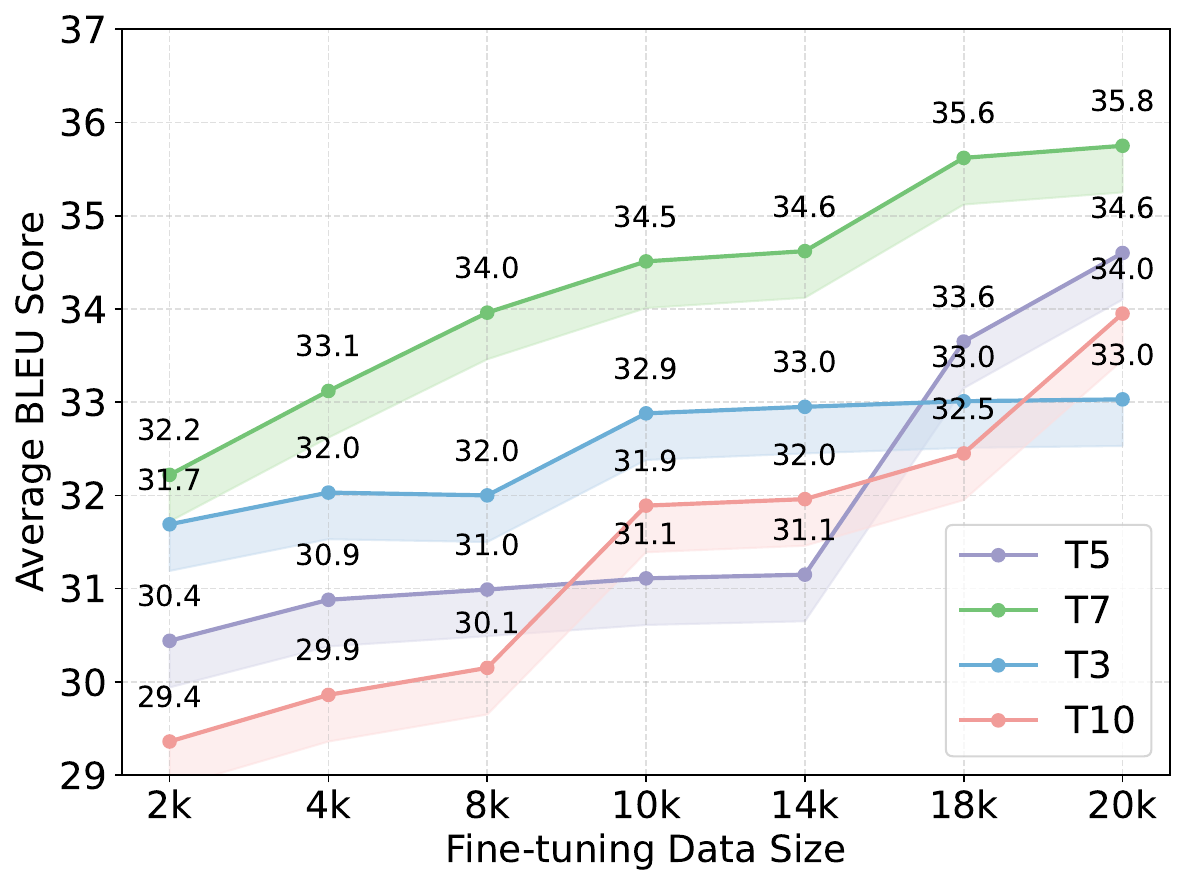}
    \caption{The average BLEU scores on the English-to-Chinese translation task with different fine-tuning data scales. The x-axis represents the amount of fine-tuning data selected from each domain.
}
    \label{tu6}
\end{figure}

Based on the experimental results in Table \ref{tablemain}, we further fine-tune the prompt strategies that benefit from domain information on Qwen-2.5-7B. The specific fine-tuned results are shown in the Figure \ref{tu6}, with the increase in fine-tuning data, all prompt strategies exhibit improved average BLEU scores. Notably, T5 shows the greatest improvement, which further highlights the effectiveness of our proposed prompt design.

\subsection{Disambiguation Performance}
As shown in the Table \ref{table5}, we further analyze the disambiguation performance of multiple disambiguation strategies, contain the following conclusion: 
\begin{tcolorbox}[colback=gray!10,colframe=black, arc=2mm,boxrule=0.5pt,width=\linewidth,boxsep=2pt, left=4pt, right=4pt,top=4pt, bottom=4pt] 
\small
\textbf{\textcolor{blue}{Conclusion 1~($\Rightarrow$ \textit{Finding 1})}}: \textbf{The disambiguation performance of the templates also improves in consistency with adding domain information.
}
\end{tcolorbox}
\begin{tcolorbox}[colback=gray!10,colframe=black, arc=2mm,boxrule=0.5pt,width=\linewidth,boxsep=2pt, left=4pt, right=4pt,top=4pt, bottom=4pt] 
\small
\textbf{\textcolor{blue}{Conclusion 2~($\Rightarrow$ \textit{Finding 2})}}: \textbf{The reason for the consistent improvement in translation performance under the reflection strategy lies in its ability to disambiguate the target translations.
}
\end{tcolorbox}

\begin{tcolorbox}[colback=gray!10,colframe=black, arc=2mm,boxrule=0.5pt,width=\linewidth,boxsep=2pt, left=4pt, right=4pt,top=4pt, bottom=4pt] 
\small
\textbf{\textcolor{blue}{Conclusion 3~($\Rightarrow$ \textit{Finding 3})}}: \textbf{In domains with stronger features (\textit{i.e., }Law and Science domains), the improvement in disambiguation accuracy is also greater. This further demonstrates that using this metric in these domains can better assess translation performance.
}
\end{tcolorbox}

\begin{tcolorbox}[colback=gray!10,colframe=black, arc=2mm,boxrule=0.5pt,width=\linewidth,boxsep=2pt, left=4pt, right=4pt,top=4pt, bottom=4pt] 
\small
\textbf{\textcolor{blue}{Conclusion 4~($\Rightarrow$ \textit{Finding 4})}}: \textbf{Disambiguation accuracy reflects improvements or declines in consistency, thereby avoiding inconsistencies in the increases or decreases of BLEU and COMET scores.
}
\end{tcolorbox}

Overall, the proposed disambiguation accuracy further demonstrates the effectiveness of the prompt strategies and corroborates the findings discussed above. The detailed disambiguation accuracy results for German-English and GPT-4o-mini are provided in Appendix \ref{appdis}, respectively.

\vspace{0.5\baselineskip}


\section{Related work}\label{s2}
\textbf{Multi-domain  Translation.}
MDT seeks to  design a unified NMT model  to translate texts
 across various domains, which can be divided into sentence-level \cite{kobus-etal-2017-domain,britz-etal-2017-effective,tars-fishel-2018-multi,aharoni-goldberg-2020-unsupervised} and word-level \cite{zeng-etal-2018-multi,8907409,jiang-etal-2020-multi-domain,lai-etal-2022-improving-domain,zhang2021domain,10328713} domain representation learning. Recently, some researchers have explored the MDT on LLMs \cite{hu-etal-2024-large-language,zheng2024fine}. 
 These methods based on conventional encoder-decoder framework.
 However, we aim to explore the performance of disambiguate when utilizing the disambiguation prompt strategies in LLMs.
 
\vspace{0.5\baselineskip}
\noindent
\textbf{Disambiguation Evaluation for Translation.} Ambiguity has long been a central challenge in machine translation, with numerous studies conducting evaluations in general domains \cite{campolungo2022dibimt,maheshwari2024dictdis,martelli2025dibimt,hu2024large}. In addition, some work has attempted to improve disambiguation by incorporating domain-specific dictionaries through constraint-based translation \cite{song2019code,chen2021lexical,zhang2023disambiguated,baek2023towards}. 
However, our approach fundamentally differs in that it does not rely on external constraint resources such as dictionaries. Instead, we focus on systematically evaluating and enhancing the disambiguation capabilities of LLMs.

\vspace{0.5\baselineskip}
\noindent
\textbf{LLMs for Translation.} These work can be broadly divided into two main categories. The first category focuses on leveraging prompting techniques \cite{vilar2022prompting,jiao2023chatgptgoodtranslatoryes,zhang2023prompting,moslem2023adaptive,he2024exploring} to enhance and analyze the performance of machine translation using LLMs. The second category focuses on fine-tuning LLMs to improve their performance in downstream NLP tasks \cite{xu2023paradigm,jiao2023parrot,zeng2023tim}. 
Our key contribution is identifying essential MDT disambiguation information for LLMs and designing prompt strategies.




\section{Conclusion}\label{s6}
In this work, we propose \textbf{DMDTEval}, a systematic benchmark for evaluating the disambiguation capabilities of LLMs in MDT. We construct a dedicated ambiguous word dataset, explore diverse prompting strategies, and evaluate five leading open-source LLMs across four language pairs and thirteen domains. Our analysis reveals key challenges in MDT disambiguation and provides actionable insights for improving domain-aware translation. In future work, we plan to develop improved methods and mechanism for disambiguation building upon the proposed ambiguous word dataset.

\section*{Limitations} 
Multi-domain bilingual parallel corpora are scarce and difficult to obtain at scale, which poses a fundamental challenge for research on MDT. Although the datasets used in our study are widely adopted in previous work and cover a broad range of domains, we acknowledge that they may contain noise, inconsistencies, or domain overlaps that could affect evaluation outcomes. Furthermore, the provenance and annotation quality of some datasets are not always transparent or verifiable, which may introduce bias into the model's disambiguation assessment.
Another limitation lies in the uneven distribution of domain data across different language pairs. While we focus on English–Chinese and German–English due to our linguistic expertise, other language pairs (\textit{e.g.,} Japanese–English and Korean–English) are only used for evaluation purposes and lack the same level of manual verification and ambiguous word coverage.
In future work, we plan to address these limitations by focusing on high-quality data collection and annotation.

\bibliography{main}

\appendix

\section{Dataset}\label{appdata}
As shown in Table \ref{t111}, the training and testing data sizes for the English-Chinese and German-English datasets are presented. Besides these two language pairs, we  use data sets from FLORES\footnote{https://huggingface.co/datasets/facebook/flores} \cite{nllb2022}, selecting Japanese-English and Korean-English. The test set consists of 1,012 sentences covering three domains: Wikinews, Wikibooks, and Wikiyago, referred to in  paper as the news, book, and travel domains. After domain-wise splitting, the data for these three domains consist of 341, 351, and 321 sentences, respectively. Additionally, regarding the scoring of alignment quality, we observe that our human annotated results achieve higher accuracy, as shown in Table \ref{table77}.
\begin{table}[t]\small
\small
\centering
\begin{tabular}{lccccc}
\toprule
\multicolumn{6}{c}{\textbf{English-Chinese}} \\
\midrule
\multicolumn{1}{l}{\textit{Train set}} & Edu & Laws & News & Sci & Spo \\
\midrule
\multicolumn{1}{l}{} & 444K & 207K & 443K & 263K & 210K \\
\midrule
\multicolumn{1}{l}{\textit{Test set}} & Edu& Laws & News & Sci & Spo \\
\midrule
\multicolumn{1}{l}{} & 790 & 456 & 1500 & 503 & 455 \\
\midrule
\midrule
\multicolumn{6}{c}{\textbf{German-English}} \\
\midrule
\multicolumn{1}{l}{\textit{Train set}} & IT & Kor & Laws & Med& Sub \\
\midrule
\multicolumn{1}{l}{} & 211K & 16K & 434K & 233K & 470K \\
\midrule
\multicolumn{1}{l}{\textit{Test set}} & IT & Kor & Laws & Med & Sub \\
\midrule
\multicolumn{1}{l}{} & 2000 & 2000 & 2000 & 2000 & 2000 \\
\bottomrule
\end{tabular}
\caption{The statistics of multi-domain translation data sets. Edu represents for the Education domain, Sci represents for the Science domain, Spo represents for the spoken domain, and Sub represents for the subtitles domain.}
\label{t111}
\end{table}
\begin{table}[]
    \centering
    \begin{tabular}{lccc}
         \toprule
\textbf{Domains}         & \textbf{C}&\textbf{Pc}&\textbf{I}\\
         \midrule
        Education&89\%&9\%&2\%\\
         Laws& 95\%&4\%&1\%\\
       News & 84\%&14\%&2\%\\
     Science & 87\%&10\%&3\%\\
     Spoken&82\%&16\%&2\%\\
         \bottomrule
    \end{tabular}
    \caption{C represents for the Correct label, Pc represents for the Partially correct label, and I represents for the Incorrect, respectively.}
    \label{table77}
\end{table}
\section{Evaluation and Training Details}\label{appbb}
 Specifically, we chose the default hyperparameter settings in vLLM for all our experiments, \textit{i.e.,} 0.8 as temperature 4, 0.95 for top\_p. The input
sequence length was chosen as 1024 for zero-shot
and CoT inference and 3000 for few-shot inference.
For the training procedure, we utilize the
LoRA \cite{hu2021lora} to fine-tune the Qwen-2.5-7B.
The hyper-parameters for supervised fine-tuning
are listed in Table \ref{table88}.

\begin{table}[t]
    \centering
    \begin{tabular}{lc}
    \toprule
      \textbf{Hyper-Parameter}   &   \textbf{Value} \\
      \midrule
         lora rank & 8\\
learning rate& 1e-5\\
train epoch& 2\\
per\_device\_batchsize &1\\
warm up ratio& 0.1\\
learning rate scheduler& cosine\\
         \bottomrule
    \end{tabular}
    \caption{Hyper-parameters for supervised fine-tuning}
    \label{table88}
\end{table}
\section{Detailed Results on LLMs}\label{appllm}
 As shown in Table \ref{t1-t4} and Table \ref{table10}, the two tables present the detailed results of our main experiments, including those from the neural machine translation model NLLB, the larger-scale model Qwen-2.5-14B, as well as the results obtained using GPT-4o-mini.

\begin{table*}[t]
\small
\centering
\begin{tabular}{llcccccc}
\toprule

&& \textbf{Education} & \textbf{Laws} & \textbf{News} & \textbf{Science} & \textbf{Spoken} & \textbf{AVG} \\
\midrule
\multicolumn{8}{c}{\textit{Neural Machine Translation-based Methods}}\\
\midrule
{\textbf{NLLB}}& / & 25.03 / 79.65 & 38.61 / 84.73 & 16.85 / 74.87 & 17.37 / 76.40 & 11.43 / 72.86 & 21.86 / 77.70 \\
\midrule
\multicolumn{8}{c}{\textit{Translation-based LLMs}}\\
\midrule
\multirow{4}{*}{\textbf{ALMA-7B}} 
&T1 & 27.86 / 86.87 & 23.35 / 88.95 & 28.57 / 84.02 & 25.39 / 84.39 & 17.59 / 76.32 & 24.75 / 84.91 \\
&T2 & 30.45 / 87.14 & 41.43 / 89.24 & 28.13 / 83.99 & 25.79 / 84.36 & 18.65 / 76.36 & 28.09 / 84.02 \\
&T3 & 29.64 / 86.86 & 43.41 / 89.54 & 27.22 / 83.65 & 26.00 / 84.51 & 18.45 / 77.02 & 28.94 / 84.72 \\
&T4 & 27.86 / 86.88 & 24.91 / 89.07 & 28.26 / 83.93 & 25.82 / 84.50 & 18.21 / 77.52 & 25.81 / 84.58 \\
\midrule
\multicolumn{8}{c}{\textit{Open-source LLMs}}\\
\midrule
\multirow{4}{*}{\textbf{LLaMA-3-8B}} 
&T1 & 22.97 / 77.40 & 22.88 / 71.30 & 16.03 / 72.31 & 15.81 / 74.04 & 15.62 / 72.46 & 18.06 / 73.30\\
&T2 & 22.70 / 79.50 & 31.21 / 73.31 & 21.32 / 74.89 & 19.87 / 76.00 & 17.10 / 72.05 & 22.04 / 75.15 \\
&T3 & 28.20 / 86.67 & 43.27 / 87.67 & 23.92 / 82.95 & 22.59 / 84.01 & 18.20 / 73.45 & 27.64 / 83.75 \\
&T4 & 20.37 / 78.76 & 26.53 / 73.23 & 17.77 / 76.09 & 17.67 / 76.69 & 15.51 / 70.32 & 19.77 / 74.82\\
\hdashline
\multirow{4}{*}{\textbf{Mistral-7B}} 
&T1 & 14.86 / 77.96 & 26.01 / 79.96 & 16.22 / 77.40 & 15.68 / 78.56 & 10.21 / 68.69 & 16.60 / 76.51 \\
&T2 & 19.04 / 81.53 & 24.10 / 79.76 & 15.71 / 77.90 & 15.09 / 80.34 & 10.08 / 68.88 & 16.80 / 77.68 \\
&T3 & 18.22 / 82.54 & 26.12 / 82.88 & 17.01 / 79.63 & 16.21 / 80.97 & 11.23 / 69.04 & 15.96 / 79.61 \\
&T4 & 10.99 / 74.26 & 7.38 / 66.07 & 6.27 / 67.15 & 7.42 / 70.10 & 8.03 / 65.11 & 7.42 / 68.14\\
\hdashline
\multirow{4}{*}{\textbf{Gemma-2-9B}} 
&T1 & 15.62 / 77.05 & 20.03 / 81.87 & 15.96 / 78.28 & 17.66 / 78.54 & 12.10 / 72.33 & 16.67 / 77.61 \\
&T2 & 16.32 / 79.09 & 20.36 / 83.23 & 16.56 / 79.51 & 18.16 / 80.83 & 12.06 / 71.03 & 17.85 / 80.67 \\
&T3 & 18.12 / 81.08 & 20.66 / 83.35 & 16.78 / 79.80 & 18.99 / 82.79 & 13.11 / 72.86 & 17.93 / 79.98 \\
&T4 & 14.69 / 71.78 & 13.16 / 69.12 & 12.33 / 70.57 & 15.25 / 71.26 & 11.10 / 66.42 & 13.31 / 69.83 \\
\hdashline
\multirow{4}{*}{\textbf{Qwen-2.5-7B}} 
&T1 & 33.14 / 88.10 & 50.82 / 88.94 & 30.04 / 84.51 & 28.76 / 84.82 & 19.20 / 77.00 & 32.39 / 84.67 \\
&T2 & 34.02 / 88.06 & 51.19 / 89.60 & 30.51 / 84.91 & 28.82 / 85.91 & 22.45 / 79.31 & 33.40 / 85.56 \\
&T3 & 34.17 / 88.17 & 50.48 / 89.22 & 29.91 / 84.66 & 28.33 / 85.64 & 18.44 / 77.12 & 32.27 / 84.96 \\
&T4 & 26.75 / 86.06 & 47.77 / 87.76 & 26.16 / 82.71 & 25.90 / 84.03 & 17.01 / 76.05 & 28.72 / 83.32 \\
\hdashline
\multirow{4}{*}{\textbf{Qwen-2.5-14B}} 
&T1 & 36.14 / 89.45 & 53.69 / 89.36 & 34.75 / 87.20 & 30.55 / 88.22 & 23.65 / 80.02 & 35.76 / 86.85
\\
&T2 & \textbf{37.90} / \textbf{89.65} & \textbf{53.87} / \textbf{89.98} & \textbf{35.14} / \textbf{87.72} & \textbf{31.04} / \textbf{88.80} & \textbf{23.51} / {80.00} & 
\textbf{36.29} / \textbf{87.23} \\
&T3 & 35.77 / 88.12 & 52.94 / 89.12 & 34.58 / 86.87 & 30.23 / 88.01 & 23.55 / 79.49 & 35.41 / 86.32\\
&T4 & 37.19 / 89.82 & 53.16 / 89.25 & 34.64 / 87.58 & 30.82 / 88.03 & 23.42 / \textbf{80.05} & 35.85 / 86.95 \\
\bottomrule
\end{tabular}
\caption{BLEU and COMET scores on the English-to-Chinese translation task (T1-T4) with different open-source LLMs and NMT models. The best results are highlighted in \textbf{bold}.}
\label{t1-t4}
\end{table*}

\begin{table*}[t]
\begin{center}
\begin{tabular}{lccccccc}
\toprule
\multicolumn{7}{c}{\textbf{English-to-Chinese}}\\
\midrule
&{\textbf{Education}}&{\textbf{Laws}}&{\textbf{News}}&{\textbf{Science}}&{\textbf{Spoken}}&{\textbf{AVG}}\\
\midrule
T1&36.77 /	88.74&50.32 / 90.31&33.00 /	85.51&30.91 /	86.61&20.25 /	79.68&34.25/	86.17\\
\hdashline
\textcolor{orange}{T5}&36.90 /	88.79&50.52 /	90.39&33.19 /	85.60&30.93 /	86.77&21.55 /	80.01&\textbf{34.62} /	\textbf{86.31}\\
\textcolor{orange}{T6}&36.58 /	88.60&50.47 /	80.42&32.55 /	85.47&30.25 /	86.39&19.50 /	79.22&33.87 /	84.02\\		
\midrule
T2			
&37.22 /	88.15&49.81 /	90.19&32.68 /	85.45&31.08 /	86.24&22.56 /	80.97&34.67 /	86.20\\
\hdashline
\textcolor{orange}{T7}&38.96 /	88.61&49.86 /	90.21&32.74 /	85.54&32.88 /	86.42&23.45 /	81.05&\textbf{35.58} /	\textbf{86.37}\\
\textcolor{orange}{T8}&34.22 /	80.23&45.36 /	85.63&30.26 /	83.00&31.11 /	85.01&22.09 /	79.65&32.61 /	82.70\\
\midrule
T3&35.15 /	88.40&52.63 /	90.54&33.12 /	85.48&30.74 /	86.59&20.13 /	79.60&34.35 /	86.12\\
\hdashline
\textcolor{orange}{T9}&
36.99 /	89.32&52.62 /	90.69&33.25 /	86.55&31.45 /	87.62&22.11 /	80.87&\textbf{35.28} /	\textbf{87.01}\\
\midrule
T4&35.97 /	88.50&50.60 /	90.14&32.74 /	85.43&31.24 /	86.76&21.69 /	80.88&34.45 /	86.34\\
\hdashline
\textcolor{orange}{T10}&
36.71 /	89.12&50.86 /	90.33&33.16 /	86.19&31.42 /	86.01&22.17 /	80.89&\textbf{34.86} /	\textbf{86.51}\\
\bottomrule
\end{tabular}
\end{center}
\caption{BLEU and COMET scores on the English-to-Chinese translation task for T1-T10 with \textbf{GPT-4o-mini}. We \textbf{bold} the best performance results in each strategy. \textcolor{orange}{Orange} text stands for the disambiguation prompt strategies.}
\label{table10}
\end{table*}

\section{Results on Other Language Pairs}\label{appyuyan}
As shown in Tables \ref{table11}, \ref{table12}, and \ref{table13}, we provide detailed experimental results for German-English, Japanese-English, and Korean-English. The overall trends are consistent with those observed in English-Chinese, further demonstrating the effectiveness of the disambiguation prompt strategies. For German-to-English, Table \ref{table11} shows that disambiguation prompting (T5–T10) consistently improves translation quality over the baseline prompts (T1–T4) across all domains. Notably, gains are most prominent in technical domains like Medical and Laws. This highlights the effectiveness of our disambiguation strategy in enhancing both BLEU and COMET scores for German-to-English translation.

\begin{table*}[t]
\begin{center}
\begin{tabular}{lccccccc}
\toprule
\multicolumn{7}{c}{\textbf{German-to-English}}\\
\midrule
&{\textbf{IT}}&{\textbf{Koran}}&{\textbf{Laws}}&{\textbf{Medical}}&{\textbf{Subtitles}}&{\textbf{AVG}}\\
\midrule
T1&29.67 / 60.08&12.85 / 69.12&25.20 / 75.69&22.06 / 72.67&20.65 / 74.20&22.09 / 70.35\\
\hdashline
\textcolor{orange}{T5}&30.37 / 72.03&15.10 / 70.26&31.05 / 80.33&32.26 / 76.98&25.42 / 74.86&\textbf{26.84 }/ \textbf{74.89}\\
\textcolor{orange}{T6}&23.54 / 67.82&14.64 / 69.79&28.42 / 78.66&32.63 / 77.63&18.74 / 70.77&{23.59} / {72.93}\\
\midrule
T2&33.56 / 80.28&15.50 / 71.38&32.98 / 82.87&36.41 / 81.64&26.40 / 76.95&\textbf{{28.97}} / \textbf{78.62}\\
\hdashline
\textcolor{orange}{T7}&33.01 / 77.45&14.34 / 70.50&29.97 / 81.38&33.47 / 79.96&25.54 / 76.62&{27.27} / {77.18}\\
\textcolor{orange}{T8}&30.91 / 72.82&14.60 / 70.32&29.59 / 79.93&31.86 / 77.69&25.05 / 76.13&26.40 / 75.38\\
\midrule
T3&33.67 / 77.75&15.61 / 71.24&33.18 / 82.61&34.58 / 79.59&25.72 / 76.52&28.55 / 77.54\\
\hdashline
\textcolor{orange}{T9}&33.34 / 77.50&15.78 / 71.46&33.90 / 82.71&34.30 / 79.61&25.82 / 76.51&\textbf{28.63 }/ \textbf{77.56}\\
\midrule
T4&28.90 / 76.01&13.52 / 69.48&29.03 / 80.50&33.01 / 79.29&21.16 / 73.77&25.13 / 75.81\\
\hdashline
\textcolor{orange}{T10}&32.14 / 77.74&15.41 / 71.15&30.97 / 82.03&33.34 / 79.45&25.48 / 75.99&\textbf{27.47} / \textbf{77.27}\\
\Xhline{1pt}
\end{tabular}
\end{center}
\caption{BLEU and COMET scores on the German-to-English translation task for T1-T10 with Qwen-2.5-7B. We \textbf{bold} the best performance results in each strategy. \textcolor{orange}{Orange} text stands for the disambiguation prompting}
\label{table11}
\end{table*}

\begin{table*}[t]
\begin{center}
\begin{tabular}{lccccc}
\toprule
\multicolumn{5}{c}{\textbf{Japanese-to-English}}\\
\midrule
&{\textbf{Book}}&{\textbf{Travel}}&{\textbf{News}}&{\textbf{AVG}}\\
\midrule
T1&23.00 / 85.38&17.47 / 84.65&21.86 / 85.97&20.78 / 85.33\\
\hdashline
\textcolor{orange}{T5}&23.66 / 85.96&17.87 / 85.02&  22.34 / 86.29& \textbf{21.29} / \textbf{85.76}\\
\textcolor{orange}{T6}&21.07 / 83.45 &16.75 / 84.01&21.43 / 85.66&19.75 / 84.37
\\		
\midrule
T2			
&24.36 / 86.36&18.50 / 85.20&23.90 / 87.49&22.25 / 86.35\\
\hdashline
\textcolor{orange}{T7}&24.71 / 86.48&18.69 / 85.48&24.23 / 87.60&\textbf{22.54} / \textbf{86.52}\\
\textcolor{orange}{T8}&22.14 / 83.75&16.85 / 84.10&20.10 / 85.01&19.70 / 84.29\\
\midrule
T3&22.96 / 84.03&16.89 / 83.99&20.65 / 83.20&20.17 / 83.74\\
\hdashline
\textcolor{orange}{T9}&23.20 / 84.76&17.56 / 84.79&21.94 / 86.01&\textbf{20.90} / \textbf{85.19}\\
\midrule
T4&24.10 / 85.96&17.97 / 85.00&21.98 / 86.01& \textbf{21.35} / \textbf{85.66}\\
\hdashline
\textcolor{orange}{T10}&22.80 / 85.10&17.12 / 84.33&21.01 / 85.46&20.31 / 84.96\\
\bottomrule
\end{tabular}
\end{center}
\caption{BLEU and COMET scores on the Japanese-to-English translation task for T1-T10 with {Qwen-2.5-7B}. We \textbf{bold} the best performance results in each strategy. \textcolor{orange}{Orange} text stands for the disambiguation prompt strategies.}
\label{table12}
\end{table*}

\begin{table*}[t]
\begin{center}
\begin{tabular}{lccccc}
\toprule
\multicolumn{5}{c}{\textbf{Korean-to-English}}\\
\midrule
&{\textbf{Book}}&{\textbf{Travel}}&{\textbf{News}}&{\textbf{AVG}}\\
\midrule
T1&23.34 / 85.55&21.22 / 85.69&21.81 / 86.25&22.12 / 85.83\\
\hdashline
\textcolor{orange}{T5}&24.05 / 86.15&22.64 / 85.78& 22.77 / 86.10 &\textbf{23.15} / \textbf{86.01}\\
\textcolor{orange}{T6}&22.98 / 84.61&20.48 / 85.14& 20.45 / 86.01&21.30 / 85.25\\		
\midrule
T2			
&23.98 / 85.70&21.58 / 85.74&22.60 / 86.43&22.72 / 85.96 \\
\hdashline
\textcolor{orange}{T7}&24.13 / 86.57&22.80 / 86.05&22.86 / 86.51&\textbf{23.26} / \textbf{86.38}\\
\textcolor{orange}{T8}&23.01 / 85.09&21.17 / 85.44&21.65 / 86.08&21.94 / 85.54\\
\midrule
T3&23.01 / 85.19 & 21.08 / 85.46&22.64 / 86.41 &22.24 / 85.69\\
\hdashline
\textcolor{orange}{T9}&23.58 / 85.67&21.85 / 85.70&22.80 / 86.57 & \textbf{22.74} / \textbf{85.98}\\
\midrule
T4&23.88 / 85.60&22.14 / 86.10&21.85 / 86.31&22.62 / 86.00\\
\hdashline
\textcolor{orange}{T10}&23.98/ 85.62&22.45 / 86.13&22.07/ 86.50&\textbf{22.83} / \textbf{86.08}\\
\bottomrule
\end{tabular}
\end{center}
\caption{BLEU and COMET scores on the Korean-to-English translation task for T1-T10 with {Qwen-2.5-7B}. We \textbf{bold} the best performance results in each strategy. \textcolor{orange}{Orange} text stands for the disambiguation prompt strategies}
\label{table13}
\end{table*}

\section{Disambiguation Performance and GPT-4o-mini Evaluator}\label{appdis}
\textbf{Disambiguation Accuracy}. As shown in Table \ref{table14}, we present the disambiguation accuracy for German-to-English, further demonstrating the robustness of our evaluation framework across multiple language pairs. This indicates that our proposed method is not limited by specific languages and can be effectively generalized to other language directions. 

\vspace{0.5\baselineskip}

\begin{figure*}
    \centering
    \includegraphics[width=\linewidth]{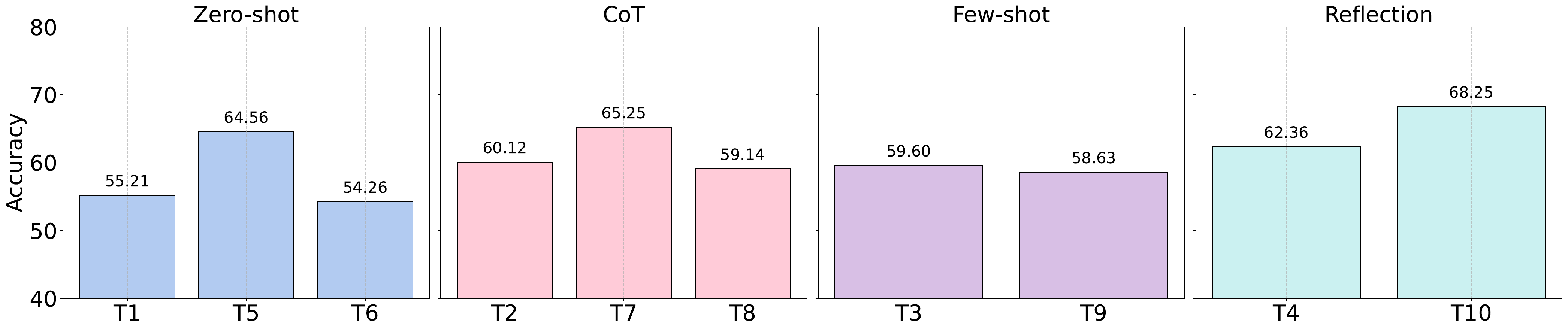}
    \caption{The accuracy of GPT-4o-mini Evaluator on the English-to-Chinese translation task.}
    \label{figure7}
\end{figure*}

\begin{figure*}
    \centering
    \includegraphics[width=\linewidth]{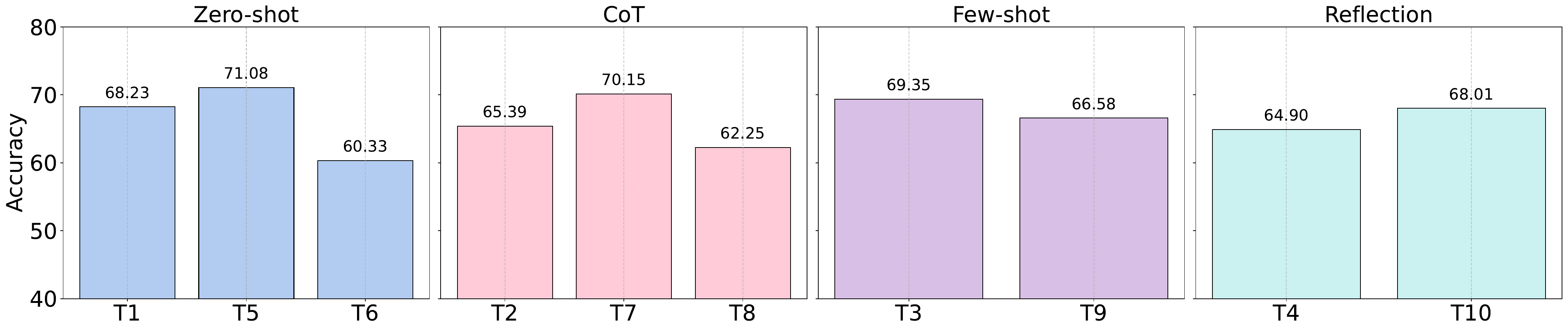}
    \caption{The accuracy of GPT-4o-mini Evaluator on the German-to-English translation task.}
    \label{figure8}
\end{figure*}

\noindent
\textbf{GPT-4o-mini Evaluator.} We design a prompt to evaluate the disambiguation capability of LLMs using GPT-4o-mini. The specific prompt is: “\textit{source sentence: < >, target sentence: < >, generate sentence: < >. Please find the ambiguous word pairs in the source language sentence and the target language sentence, and count the number of ambiguous word pairs. Refer to the above word pairs to further count the accuracy of disambiguation in the generated sentences.} ”. 
We calculate the average accuracy across different templates with GPT-4o-mini in  Figure \ref{figure7} and \ref{figure8}. The consistency with Figure \ref{figure7} and \ref{figure8} further prove the effectiveness of disambiguation prompting. 

\begin{table*}[t]
   \begin{center}
\begin{tabular}{lclllllll}
\toprule
\multirow{2}{*}{\textbf{Strategies}}&\multicolumn{6}{c}{\textbf{German-to-English}}&\multirow{2}{*}{{\textbf{AVG}}}\\
\cmidrule{3-7}
&&{\textbf{IT}}&{\textbf{Koran}}&{\textbf{Laws}}&{\textbf{Medical}}&{\textbf{Subtitles}}&\\
\midrule
\multirow{3}{*}{\textbf{\makecell[l]{Zero-shot}}}&T1 & 51.87 & 53.42 & 59.11 & 50.93 & 55.07 & 54.08 \\
&\textcolor{orange}{T5} 
& 58.26\textsubscript{\textcolor{blue}{+6.39}} 
& 57.89\textsubscript{\textcolor{blue}{+4.47}} 
& 60.12\textsubscript{\textcolor{blue}{+1.01}} 
& 57.34\textsubscript{\textcolor{blue}{+6.41}} 
& 58.90\textsubscript{\textcolor{blue}{+3.83}} 
& \textbf{58.50}\textsubscript{\textcolor{blue}{+4.42}} \\
&\textcolor{orange}{T6}
&48.62\textsubscript{\textcolor{red}{-3.25}}
&52.15\textsubscript{\textcolor{red}{-5.27}}
&56.49\textsubscript{\textcolor{red}{-3.63}}
&48.80\textsubscript{\textcolor{red}{-8.54}}
&51.23\textsubscript{\textcolor{red}{-7.67}}
&51.46\textsubscript{\textcolor{red}{-2.62}} \\
\hdashline
\multirow{3}{*}{\textbf{\makecell[l]{CoT}}}&T2&55.89&58.07&60.35&56.88&57.31&57.70\\
&\textcolor{orange}{T7}
&58.02\textsubscript{\textcolor{blue}{+2.13}}
&59.86\textsubscript{\textcolor{blue}{+1.79}}
&61.17\textsubscript{\textcolor{blue}{+0.82}}
&58.02\textsubscript{\textcolor{blue}{+1.14}}
&58.67\textsubscript{\textcolor{blue}{+1.36}}
&\textbf{59.15}\textsubscript{\textcolor{blue}{+1.45}} \\
&\textcolor{orange}{T8}
&44.99\textsubscript{\textcolor{red}{-13.03}}
&48.25\textsubscript{\textcolor{red}{-10.62}}
&47.15\textsubscript{\textcolor{red}{-14.02}}
&43.08\textsubscript{\textcolor{red}{-14.94}}
&49.19\textsubscript{\textcolor{red}{-9.48}}
&46.53\textsubscript{\textcolor{red}{-11.17}} \\
\hdashline
\multirow{2}{*}{\textbf{\makecell[l]{Few-shot}}}
&T3 & 53.10 & 54.97 & 60.45 & 51.65 & 55.81 & \textbf{55.20} \\
&\textcolor{orange}{T9} 
&46.23\textsubscript{\textcolor{red}{-6.87}} 
&50.78\textsubscript{\textcolor{red}{-4.19}} 
&53.62\textsubscript{\textcolor{red}{-6.83}} 
&48.84\textsubscript{\textcolor{red}{-2.81}} 
&51.87\textsubscript{\textcolor{red}{-3.94}} 
&50.27\textsubscript{\textcolor{red}{-4.93}} \\
\hdashline
\multirow{2}{*}{\textbf{\makecell[l]{Reflection}}}
&T4 & 56.92 & 53.75 & 58.90 & 52.41 & 56.30 & 55.46 \\
&\textcolor{orange}{T10} 
&60.02\textsubscript{\textcolor{blue}{+3.10}} 
&57.81\textsubscript{\textcolor{blue}{+4.06}} 
&59.51\textsubscript{\textcolor{blue}{+0.61}} 
&55.89\textsubscript{\textcolor{blue}{+3.48}} 
&59.48\textsubscript{\textcolor{blue}{+3.18}} 
&\textbf{58.54}\textsubscript{\textcolor{blue}{+3.08}} \\
\bottomrule
\end{tabular}
\end{center}
\caption{Disambiguation accuracy scores (\%) on the German-to-English translation task for T1-T10 with Qwen-2.5-7B. We \textbf{bold} the best performance results in each strategy.}
\label{table14}
\end{table*}

\end{CJK*}
\end{document}